\documentclass{egpubl}
\usepackage{pg2025}

\SpecialIssuePaper         

\CGFccby

\usepackage[T1]{fontenc}
\usepackage{dfadobe}  

\usepackage{cite}  
\BibtexOrBiblatex
\electronicVersion
\PrintedOrElectronic
\ifpdf \usepackage[pdftex]{graphicx} \pdfcompresslevel=9
\else \usepackage[dvips]{graphicx} \fi

\usepackage{egweblnk} 

\title[Preconditioned Deformation Grids]%
      {Preconditioned Deformation Grids}

\author[J. Kaltheuner \& A. Oebel \& H. Droege \& P. Stotko \& R. Klein]
{\parbox{\textwidth}{\centering} {Julian Kaltheuner$^{1}$\orcid{0000-0002-5218-1638},
        Alexander Oebel$^{1}$\orcid{0000-0002-4039-2438},
        Hannah Droege$^{1}$\orcid{0000-0001-7163-4279},
        Patrick Stotko$^{1}$\orcid{0000-0002-2608-0278},
        Reinhard Klein$^{1}$\orcid{0000-0002-5505-9347}
        }
        \\
{\parbox{\textwidth}{\centering $^1$University of Bonn, Germany\\
       }
}
}

\usepackage{amsmath}
\usepackage{amssymb}

\usepackage{cleveref}
\usepackage{booktabs}
\usepackage{multirow}

\Crefname{figure}{Fig.}{Figs.}
\Crefname{section}{Sec.}{Secs.}

\usepackage{pifont}
\newcommand{\cmark}{\ding{51}}%
\newcommand{\xmark}{\ding{55}}%

\usepackage{xspace}
\def\etal{et al.\xspace}

\newcommand{\abs}[1]{\left\vert#1\right\vert}
\newcommand{\norm}[1]{\left\Vert#1\right\Vert}

\usepackage{bm}

\renewcommand{\vec}[1]{\bm{#1}}
\newcommand{\mat}[1]{\bm{#1}}
\newcommand{\set}[1]{\mathcal{#1}}

\DeclareMathOperator*{\argmin}{arg\,min}
\DeclareMathOperator*{\argmax}{arg\,max}
\DeclareMathOperator*{\sg}{sg}

\let\originalleft\left
\let\originalright\right
\renewcommand{\left}{\mathopen{}\mathclose\bgroup\originalleft}
\renewcommand{\right}{\aftergroup\egroup\originalright}

\begin{document}

\teaser{
    \includegraphics[width=1\linewidth]{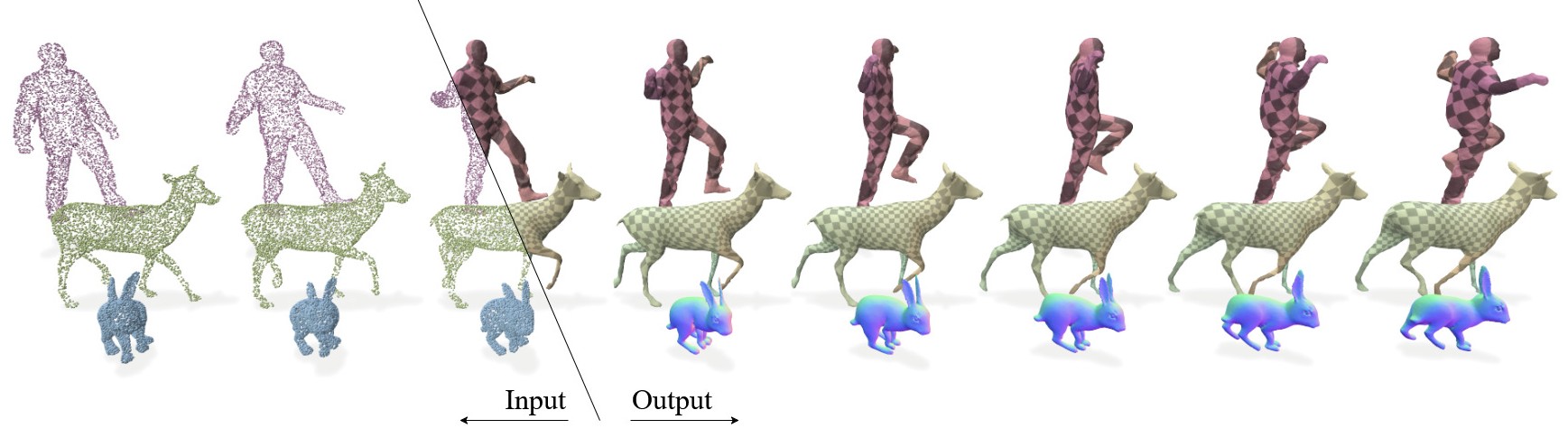}
    \centering
    \caption{Raw point cloud sequences (left) are unstructured and lack temporal correspondences.
    We propose Preconditioned Deformation Grids (right), a method for temporally coherent, high‐fidelity surface reconstructions that evolve smoothly over time.}
    \label{fig:teaser}
}

\maketitle
\begin{abstract}
Dynamic surface reconstruction of objects from point cloud sequences is a challenging field in computer graphics.
Existing approaches either require multiple regularization terms or extensive training data which, however, lead to compromises in reconstruction accuracy as well as over-smoothing or poor generalization to unseen objects and motions.
To address these limitations, we introduce \emph{Preconditioned Deformation Grids}, a novel technique for estimating coherent deformation fields directly from unstructured point cloud sequences without requiring or forming explicit correspondences.
Key to our approach is the use of multi-resolution voxel grids that capture the overall motion at varying spatial scales, enabling a more flexible deformation representation.
In conjunction with incorporating grid-based Sobolev preconditioning into gradient-based optimization, we show that applying a Chamfer loss between the input point clouds as well as to an evolving template mesh is sufficient to obtain accurate deformations.
To ensure temporal consistency along the object surface, we include a weak isometry loss on mesh edges which complements the main objective without constraining deformation fidelity.
Extensive evaluations demonstrate that our method achieves superior results, particularly for long sequences, compared to state-of-the-art techniques.


\end{abstract}

\section{Introduction}
\label{sec:intro}

Reconstructing dynamic 3D surfaces from temporal point cloud sequences is a fundamental problem in computer graphics with diverse applications in animation, virtual production, medical imaging, robotics, autonomous driving, as well as augmented reality (AR) and virtual reality (VR).
The increasing availability of commodity depth sensing, ranging from smartphone LiDAR to multi-camera RGB systems, has made it feasible to capture dense 3D point clouds at video rates.
However, these point clouds are typically unstructured, lack temporal correspondences, and exhibit noise or incompleteness, posing significant challenges for reliable surface reconstruction.
Naively applying static reconstruction methods~\cite{huang2023neural, peng2021shape, hanocka2020point2mesh, williams2019deep} to each frame independently fails to exploit temporal coherence, often resulting in inconsistent geometry, lost correspondences, and high computational cost.
Thus, dynamic reconstruction methods aim to estimate a consistent 3D surface for a reference frame and deform it to match subsequent points measurements across time.
Yet, this problem is inherently under-constrained, and in the absence of strong temporal cues, deformation fields may overfit the data while producing implausible motion.

To mitigate these issues, previous approaches incorporate temporal priors to enforce continuity across frames.
Template-based models such as SMPL~\cite{loper2015smpl} and SCAPE~\cite{anguelov2005scape} encode correspondences via parametric shape spaces, while alternative methods estimate inter-frame deformations through learning-based or optimization-driven techniques~\cite{bozic2021neural, lei2022cadex}.
Although effective in specific domains, these approaches often rely on category-specific assumptions or extensive training data, limiting their ability to generalize to unseen object classes or non-rigid, unpredictable motions.
In addition, many methods employ strong regularization to stabilize reconstructions and enforce smoothness which often comes at the cost of geometric detail~\cite{wu2019global, yang2019global}.
This trade-off between fidelity and stability can undermine subtle but meaningful surface variations, motivating the need for more flexible formulations.
To overcome these limitations, it is essential to develop a framework that jointly exploits temporal coherence without relying on restrictive priors, while still preserving high‐frequency details across challenging dynamic sequences.

In this work, we introduce \textit{Preconditioned Deformation Grids}, a correspondence-free, training-free framework for reconstructing temporally coherent, high-fidelity surfaces directly from unstructured point cloud sequences.
Our method begins by selecting a suitable keyframe to estimate an initial deformable template mesh, which is further refined as part of the optimization process.
Rather than relying on explicit correspondences or a set of carefully hand-tuned regularization terms, our approach employs Sobolev preconditioning, which spatially diffuses the under-constrained gradient information from the raw point clouds across local neighborhoods.
This effectively acts as a spatially adaptive smoothness constraint, allowing the optimization to prioritize plausible deformations without explicitly imposing any restrictions to the reconstruction fidelity.
To further guide the process, we represent the overall motion at various spatial scales using multi-resolution voxel grids.
Coarser grid levels represent broad movements and thus help maintain temporal coherence over long sequences, while finer levels enable the recovery of high-frequency surface details by allowing localized adjustments.
Thanks to the stabilizing effect of Sobolev preconditioning and the multi-scale representation of the deformation field, a simple Chamfer loss defined on the unstructured input point cloud together with a weak isometry loss on the edges of the evolving template mesh is sufficient to achieve superior results over current state-of-the-art approaches.

In summary, our main contributions are:

\begin{itemize}
    \item A correspondence‐free deformation framework that operates directly on unstructured point clouds, eliminating the need for pre-defined template models or category‐specific priors, and enables robust, large‐scale motion estimation for arbitrary objects.
    \item A multi-resolution voxel grid representation for the deformation field that models the motion at varying scales.
    \item A grid-based Sobolev preconditioning scheme that stabilizes the optimization by diffusing under-constrained gradients across neighboring grid cells.
\end{itemize}

The code of our work is available at \url{https://github.com/vc-bonn/preconditioned-deformation-grids}.

\section{Related Work}
\label{sec:relatedwork}
\subsection{Parametric Template Models}

A widely adopted strategy for category-specific 3D reconstruction involves deforming a predefined mesh template to align with observed data.
In facial reconstruction, the FLAME~\cite{li2017learning} and FaceVerse~\cite{wang2022faceverse} models define parametric spaces over shape and expression.
For full-body reconstruction, approaches like SMPL~\cite{loper2015smpl}, \mbox{SMPL-X}~\cite{pavlakos2019expressive}, SCAPE~\cite{anguelov2005scape}, as well as the more recent GHUM/GHUML~\cite{xu2020ghum} similarly encode pose and identity within low-dimensional manifolds and use corrective blend shapes and linear blend skinning to deform the template.
These models have been successfully applied to detailed human performance capture from real-world recordings~\cite{bogo2016keep, sanyal2019learning, liu2021neural}, and further extended to incorporate clothing and apparel variations~\cite{bhatnagar2019multi, alldieck2019learning, tiwari2020sizer}.

While template-based approaches offer compact, interpretable control over shape and articulation, they are fundamentally limited by the expressiveness of the underlying parametric model, which can restrict generalization to novel shapes, poses, or deformable objects outside the training distribution.

\subsection{Deformation Field Estimation}

Estimating non-rigid motion from image or point cloud data is commonly addressed through either optimization-based or learning-based methods that solve for deformation fields.
Early optimization-driven techniques model deformation directly from input data.
Deformation graphs, introduced by Sumner \etal~\cite{sumner2007embedded}, provide a sparse and flexible representation for surface motion.
This concept was extended in DynamicFusion~\cite{newcombe2015dynamicfusion}, which estimates a volumetric 6D warp field to incrementally align a canonical surface with each new input frame.
Recently, the success of 3D Gaussian Splatting (3DGS)~\cite{kerbl20233d} in efficiently representing and rendering radiance fields led to further advances in dynamic reconstruction.
For instance, 4DTAM~\cite{matsuki20254dtam} performed online 4D tracking and mapping from a single RGB-D stream using dynamic surface Gaussians, jointly optimizing geometry, appearance, camera ego-motion, and a learned warp field.
Other Gaussian-Splatting–based dynamic SLAM systems distinguished between static and dynamic scene content, either by separating them into individual Gaussian maps~\cite{li2025dynagslamrealtimegaussiansplattingslam} or by predicting uncertainty images for guiding rigid bundle andjustment~\cite{zheng2025wildgsslammonoculargaussiansplatting}.

More recently, learning-based methods have gained traction.
Neural Deformation Graphs~\cite{bozic2021neural} and Neural Non-Rigid Tracking~\cite{bozic2020neural} learn non-rigid motion by estimating point correspondences and deformation fields through neural networks.
Hierarchical models such as Neural Deformation Pyramid~\cite{li2022non} represent motion at multiple spatial scales, where each level is handled by a lightweight MLP.
An alternative line of work formulates deformation as a learned, continuous spatio-temporal vector field~\cite{niemeyer2019occupancy}, integrated via Neural ODE solvers~\cite{chen2018neural}, or using them to update latent codes that control global shapes \cite{jiang2021learning} or for mesh deformations \cite{huang2020meshode}, whereas other approaches \cite{tang2021learning} predict deformations directly.
Complementary to these vector-field formulations, Dynamic Neural Surfaces~\cite{nizamani2025dynamicneuralsurfaceselastic} represent deforming 4D shapes as continuous elastic surfaces in space–time, enabling spatio-temporal registration and statistical analysis for genus-zero surfaces.
Recent generative approaches further explore probabilistic modeling of deformation.
Motion2VecSets~\cite{cao2024motion2vecsets} employs a 4D diffusion model to learn distributions over shapes and their deformations.
Other methods aim to jointly learn a canonical shape and its deformation mappings, either via latent embeddings~\cite{lei2022cadex, yenamandra2021i3dmm} or within volumetric rendering frameworks~\cite{pumarola2021d, park2021nerfies, li2021neural}, where deformation is optimized alongside radiance fields.
In contrast to models that rely on shape priors, DynoSurf~\cite{yao2024dynosurf} introduces an unsupervised method that jointly estimates a deforming surface and template geometry directly from point cloud sequences.

In the same spirit, our method operates in a category-agnostic, training-free regime, estimating deformation fields without relying on strong priors, supervision, or correspondence annotations.

\subsection{Preconditioning}

Preconditioning is a classical strategy in inverse problems used to accelerate convergence and enhance optimization stability by carefully adapting update steps.
In geometry processing, preconditioners have been employed to improve mesh parameterization~\cite{claici2017isometry} and accelerate mesh deformation~\cite{kovalsky2016accelerated}.
Krishna \etal~\cite{krishnan2013efficient} demonstrate the effectiveness of multi-level preconditioning on Laplacian matrices, showing broad applicability across mesh-based tasks.
Despite these advances, ill-conditioned optimization problems remain a challenge in high-dimensional and irregular domains.
Recent works explore learning-based alternatives, using graph neural networks to learn effective preconditioners from data~\cite{rudikov2024neural, li2023learning, chen2024graph, hausner2024neural, trifonov2024learning}.

A particularly relevant technique in this context is Sobolev preconditioning, which replaces the standard $ L^2 $ inner product with a Sobolev (e.g., $ H^1 $) inner product to compute smoother gradient directions~\cite{neuberger1985steepest}.
Sobolev gradient methods have been widely studied for applications in surface smoothing and minimal surface flows~\cite{renka2004constructing, renka1995minimal, eckstein2007generalized}.
Martin \etal~\cite{martin2013efficient} further demonstrated the benefits of Sobolev preconditioners in the area of geometry processing and optimized shapes for smooth surfaces.
The technique has since been applied to non-rigid scene reconstruction from RGB-D data~\cite{slavcheva2018sobolevfusion}, differentiable rendering pipelines~\cite{nicolet2021large}, and registration of deforming objects~\cite{jung2025preconditioned}.
Recently, Chang \etal~\cite{chang2024spatiotemporal} proposed a variant that uses spatiotemporal bilateral gradient filtering, which diffuses gradient information for stability while preserving high-frequency details.

In this work, we adopt a grid-based Sobolev preconditioning scheme to regularize gradient updates from unstructured point cloud sequences, promoting smooth yet flexible deformation fields that improve spatial coherence without sacrificing geometric detail.

\section{Method}

\begin{figure*}
    
  \centering
  \includegraphics[width=\textwidth]{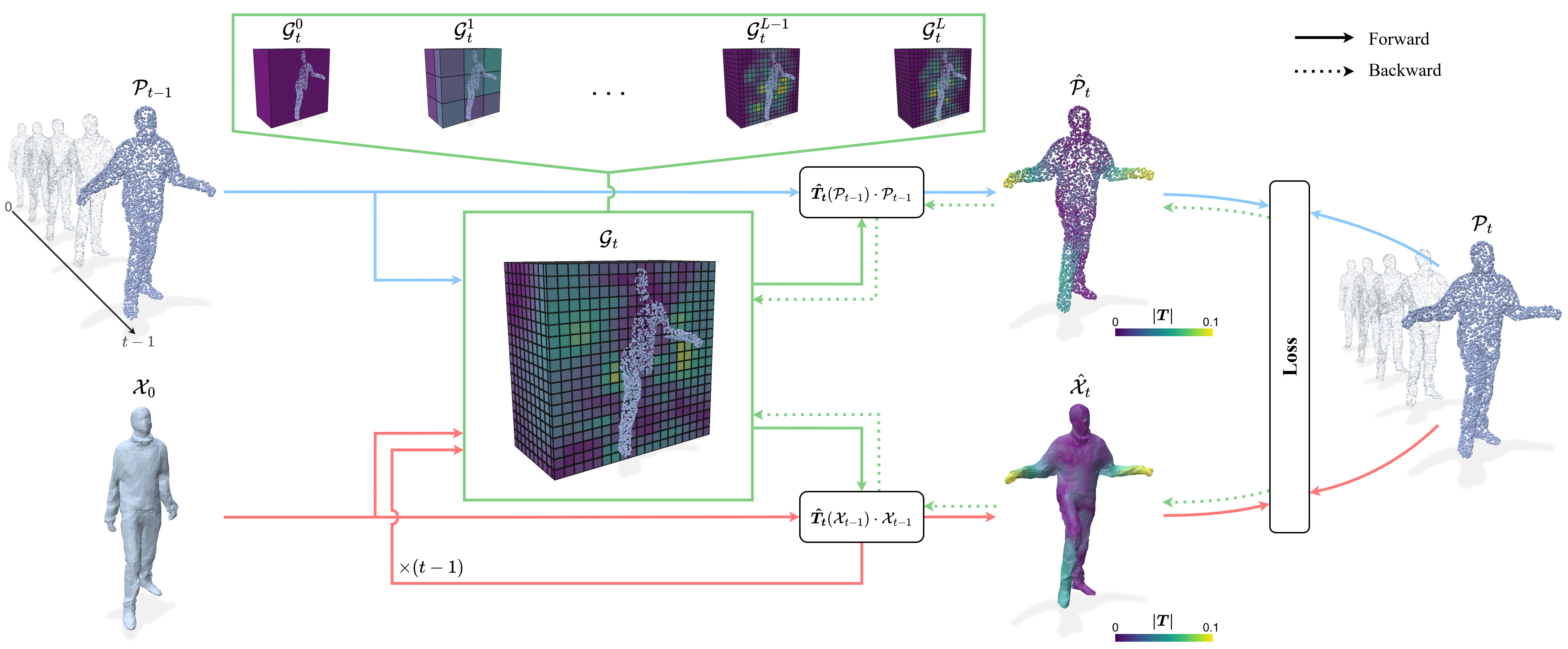}
  \caption{Overview of our correspondence-free deformation framework.
  At the heart of our approach is a multi-resolution voxel grid $ \set{G}_t $ that encodes 6D transformations at multiple spatial scales $ l $.
  Given a sequence of unstructured point clouds $ \set{P}_t $ and an estimated initial template mesh $ \set{X}_0 $, we apply grid-based preconditioning to diffuse the under-constrained gradient information during optimization of the transformations $ \vec{T}_t^{l,c} $ within the grid.}
  \label{fig:overview}
\end{figure*}

We present an optimization-based method for estimating dense surface deformations from unstructured point cloud sequences, without relying on temporal correspondences, learned priors, or pretraining.
Given a sequence of point clouds $ \{ \set{P}_t \}_{t = 0}^T $, where each $ { \set{P}_t = \{ \vec{p}_{i,t} \in \mathbb{R}^3 \} } $, and an initial mesh $ { \set{X}_0 = \{ \vec{x}_{i,0} \in \mathbb{R}^3 \} } $ defined at a keyframe (see \Cref{sec:keyframe}), our goal is to estimate a sequence of deformation fields that, when applied to $ \set{X}_0 $, yield temporally consistent surface reconstructions $ \{ \hat{\set{X}}_t \}_{t = 1}^T $.
As illustrated in \Cref{fig:overview}, we represent the deformation field as a multi-resolution voxel grid which encodes local rigid transformations at different spatial scales (see \Cref{sec:grid}).
A key technical contribution is a spatially-aware preconditioning scheme (see \Cref{sec:preconditioning}) that stabilizes the optimization by enforcing spatial smoothness in the deformation field.

\subsection{Multi-Resolution Transformation Grid}
\label{sec:grid}

To enable the estimation of temporally coherent deformations, we represent the motion from time step $ t - 1 $ to $ t $ at different spatial scales using a multi-resolution voxel grid $ { \set{G}_t = \{ G_t^l \}_{l = 1}^L } $ composed of $ L $ hierarchical levels.
Each level $ G^l $ consists of a finite set of grid cells $ { \set{C}^l \subseteq \{ c \in [0 : 2 l - 1)^3 \} } $, where each cell $ c $ stores a 6D transformation vector $ { \vec{T}_t^{l,c} \in \mathbb{R}^6 } $, parameterizing a local rigid motion via rotation and translation (see \Cref{sec:transform_param}).
In particular, the spatial resolution of the grid increases linearly by $ 2 l - 1 $ with level index $ l $, ranging a coarse global deformation at $ { l = 1 } $ to fine-scale local displacements at the finest level $ { L = 10 } $.
For an input point $ { \vec{x} \in \mathbb{R}^3 } $, the corresponding transformation $ { \hat{\vec{T}}_t(\vec{x}) \in \mathbb{R}^6 } $ is computed by aggregating the trilinear-interpolated partial transformations $ \vec{T}_t^{l,c} $ from neighboring cells $ { c \in \mathcal{N}_t^l(\vec{x}) } $ across all resolution levels $ l $:
\begin{equation}
	\hat{\vec{T}}_t(\vec{x}) = \frac{1}{L} \sum_{l = 1}^L \sum_{c \in \mathcal{N}_t^l(\vec{x})} w_t^{l,c}(\vec{x}) \cdot \vec{T}_t^{l,c},
\end{equation}
where $ w^{l,c}_{t} $ are the trilinear interpolation weights.
Our approach constructs the deformation field using a redundant, multi-scale representation, achieved by explicitly averaging transformations derived from parameters at each resolution level.
This design allows parameters at each level to model aspects of the \textit{absolute deformation} relevant to their respective scale, rather than encoding explicit \textit{residuals} of coarser levels.
The inherent redundancy in an averaged representation is crucial for enhancing optimization stability, as it provides robustness against the high-frequency variations and estimation sensitivities often encountered when learning direct residuals, which can be challenging even for preconditioned solvers.

\subsection{Spatial Smoothness via Grid Preconditioning}
\label{sec:preconditioning}
Real-world deformations exhibit strong spatial coherence: neighboring regions typically undergo similar transformations.
In the absence of regularization, the deformation estimation problem is severely under-constrained since the input points $ \set{P}_t $ provide only sparse observations, allowing many plausible interpolated deformations between observed points.
Naively optimizing transformations for each voxel independently can result in folding, tearing, or discontinuous motion fields that violate physical plausibility.

To address this, rather than imposing spatial coherence through an explicit penalty term, we incorporate smoothness directly into the optimization dynamics via preconditioning.
Let $ \vec{T}^l $ denote the stacked transformation parameters for all grid cells at resolution level $ l $.
A standard gradient descent update is given by:
\begin{equation}
	\vec{T}^l \leftarrow \vec{T}^l - \eta \, \frac{\partial \set{L}}{\partial \vec{T}^l},
\end{equation}
where $ \eta $ is the learning rate and $ \set{L} $ the loss function.
Instead, we apply spatially-aware preconditioning~\cite{nicolet2021large} that couples updates between neighboring grid cells:
\begin{equation}
	\vec{T}^l \leftarrow \vec{T}^l - \eta \, (\mathbf{I} + \lambda \, \mat{L}^l)^{-2} \, \frac{\partial \set{L}}{\partial \vec{T}^l},
\end{equation}
where $ \mat{L}^l $ is the Laplacian matrix encoding adjacency between neighboring grid cells, and $ { \lambda > 0 } $ controls the strength of spatial smoothing.
This corresponds to a heat diffusion process on the grid where, in this case, gradients between neighboring cells are continuously smoothed over time and $ \lambda $ determines the time scale of diffusion and thus the degree of smoothing.

This preconditioning approach offers several advantages over traditional energy-based regularization, as 1) the \textit{elimination of the accuracy tradeoff} inherent in explicit regularization formulations, allowing the transformations to fit the data while still enforcing smoothness.
By embedding smoothness in the optimization trajectory (rather than the final objective) serves as 2) \textit{adaptive regularization} and preserves the ability to represent non-smooth deformations, such as sharp boundaries, whenever the data demand them. 
Furthermore, in under-determined problems, where multiple solutions exist that fit the data equally well, the optimization naturally leads toward 3) \textit{smooth solutions}, thereby resolving ambiguities in a principled manner. 
Additionally, it improves 4) \textit{numerical stability} by dampening high-frequency oscillations, leading to more robust convergence compared to standard gradient descent. 
Finally, because the grid topology is fixed, the Laplacian matrix $ \mat{L}^l $ can be precomputed once, and its sparse structure exploited by standard linear solvers, leading to 5) an \textit{efficient computation}.

Note that we apply preconditioning not only to our grids, but also to the optimization of the mesh vertices $ \set{X}_0 $ (see \Cref{sec:optimization}), following the mesh preconditioning strategy by Nicolet \etal~\cite{nicolet2021large}.

\subsection{Transformation Parameterization}
\label{sec:transform_param}

As mentioned in \Cref{sec:grid}, each grid cell encodes a 6D transformation vector:
\begin{equation}
    \vec{T}_t^{l,c} = [\vec{z}^T, \vec{t}^T]^T
\end{equation}
where $ { \vec{z} = (z_0, z_1, z_2)^T \in \mathbb{R}^3 } $ parameterizes the rotation component and $ { \vec{t} = (t_x, t_y, t_z)^T \in \mathbb{R}^3 } $ specifies the translation component.
To avoid issues such as gimbal lock and to ensure robust optimization, we adopt the Cayley parameterization~\cite{zhang2021fast} to express rotations.
Specifically, the rotation matrix $ { \mat{R} \in \mathbb{R}^{3 \times 3} } $ is defined as:
\begin{equation}
	\mat{R}(\vec{z}) = (\mathbf{I} + \mat{Z}) \, (\mathbf{I} - \mat{Z})^{-1}
\end{equation}
where $ \mat{Z} $ is the skew-symmetric matrix constructed from $ \vec{z} $:
\begin{equation}
	\mat{Z}
    =
    \begin{pmatrix}
		0 & -z_2 & z_1 \\
		z_2 & 0 & -z_0 \\
		-z_1 & z_0 & 0
	\end{pmatrix}
\end{equation}
The full transformation at a point is represented by a $ { 4 \times 4 } $ homogeneous matrix, composed of the interpolated rotation $ \mat{R}(\vec{z}) $ and translation $ \vec{t} $:
\begin{equation}
	\mat{M}(\hat{\vec{T}}_t(\vec{p}))
    =
    \begin{bmatrix}
		\mat{R}(\vec{z}) & \vec{t} \\
		\vec{0}^T & 1
	\end{bmatrix}
\end{equation}
To estimate the deformed surface $ \hat{\set{X}}_t $ at time $ t $, we apply the estimated transformations recursively from $ { t = 0 } $ up to the current step to the initial mesh $ \set{X}_0 $:
\begin{equation}
	\hat{\set{X}}_t = \left\{ \hat{\vec{x}}_t \in \mathbb{R}^3 \,\middle\vert\, \hat{\vec{x}}_t = \prod_{\tau = 1}^{t} \mat{M}(\hat{\vec{T}}_{\tau}(\vec{x}_{\tau - 1})) \cdot \vec{x}_0 \right\} 
\end{equation}
For brevity, we omit the explicit conversion to and from homogeneous coordinates.

While the formulation above propagates transformations forward in time, that is from $ \hat{\set{X}}_{t - 1} $ to $ \hat{\set{X}}_t $, the same mechanism can be applied in reverse.
This allows more a flexible starting point by selecting a suitable keyframe for $ \set{X}_0 $ at any time step $ { t > 0 } $ of the input point cloud sequence (see \Cref{sec:keyframe}).

\subsection{Optimization Objectives}
\label{sec:optimization}

Our method jointly optimizes the initial surface mesh $ \set{X}_0 $ and the multi-resolution transformation grids $ \{ \vec{T}_t^{l,c} \} $ by minimizing:
\begin{equation}
	\set{L} = \set{L}_{\mathrm{mesh}} + \set{L}_{\mathrm{transform}} + w_{\mathrm{isometry}} \cdot \set{L}_{\mathrm{isometry}},
\end{equation}
where $ { w_{\mathrm{isometry}} = 250 } $ controls the contribution of the isometry loss $ \set{L}_{\mathrm{isometry}} $.
This value is chosen such that the isometry loss contributes only weakly, accounting for approximately 25\% of the typical magnitude of the transformation loss $ \mathcal{L}_{\mathrm{transform}} $.

\paragraph*{Surface Initialization Loss.}

To ensure accurate surface geometry at the keyframe, we align the initially estimated mesh $ \set{X}_0 $ to the corresponding reference point cloud $ \set{P}_0 $ by minimizing a robust variant of the Chamfer distance:
\begin{equation}
    \mathcal{L}_{\mathrm{mesh}} = \mathrm{CD}_{\mathrm{R}}(\set{X}_0, \set{P}_0),
\end{equation}
where $ \mathrm{CD}_{\mathrm{R}} $ denotes the robust Chamfer distance~\cite{yao2024dynosurf}, which is designed to reduce sensitivity to outliers:
\begin{align}
    \mathrm{CD}_{\mathrm{R}}(\set{P}, \set{Q})
    = \ &
    \frac{1}{|\set{P}|} \sum_{\vec{p} \in \set{P}} w_{\mathrm{R}}(\vec{p}, \vec{q}_{\vec{p}}) \norm{\vec{p} - \vec{q}_{\vec{p}}}^2
    \nonumber \\
    & +
    \frac{1}{|\set{Q}|} \sum_{\vec{q} \in \set{Q}} w_{\mathrm{R}}(\vec{p}_{\vec{q}}, \vec{q}) \norm{\vec{p}_{\vec{q}} - \vec{q}}^2
\end{align}
where
\begin{equation}
    \vec{q}_{\vec{p}} = \argmin_{\vec{q} \in \set{Q}} \norm{\vec{p} - \vec{q}}
    \quad , \quad
    \vec{p}_{\vec{q}} = \argmin_{\vec{p} \in \set{P}} \norm{\vec{p} - \vec{q}}
\end{equation}
are the nearest neighbors of the points $ \vec{p} $ and $ \vec{q} $ in the opposite point sets $ \set{Q} $ and $ \set{P} $, and
\begin{equation}
    w_{\mathrm{R}}(\vec{p}, \vec{q}) = \exp(-\alpha \cdot \norm{\vec{p} - \vec{q}}^2)
\end{equation}
is a robust weighting function $ w_{\mathrm{R}}(\vec{p}, \vec{q}) $ defined as a Gaussian kernel with $ { \alpha = 5.56 } $.
This formulation attenuates the influence of outliers by down-weighting correspondences with large residuals, thereby enhancing the robustness of the surface alignment.

\paragraph*{Transformation Fitting Loss.}

The primary data fitting objective encourages correct alignment between the transformed surface and the target point clouds:
\begin{equation}
	\mathcal{L}_{\mathrm{transform}} = \frac{1}{T} \sum_{t=1}^T w_{\mathrm{confidence}}(t) \cdot \mathrm{CD}_{\mathrm{R}}(\hat{\set{X}}_t, \set{P}_t) + \mathrm{CD}_{\mathrm{R}}(\hat{\set{P}}_t, \set{P}_t) ,
\end{equation}
where $ \hat{\set{P}}_t $ denotes the input points $ \set{P}_{t - 1} $ from time $ t - 1 $ transformed forward by the transformation grid $ \set{G}_t $, and $ w_{\mathrm{confidence}}(t) $ is an adaptive confidence weight.
This formulation leverages the grid structure to gather information from the input points at all timesteps, thereby guiding the mesh transformation process and enabling the estimation of robust, temporally coherent deformations.

\paragraph*{Confidence-Based Error Control.}

Due to the sequential nature of our deformation model, inaccuracies in early transformations can propagate through time, forcing subsequent steps to compensate for accumulated errors.
This often leads to increasingly large and unstable transformations, ultimately degrading reconstruction quality.
To mitigate this, we introduce an adaptive confidence weighting scheme:
\begin{equation}
	w_{\mathrm{confidence}}(t) = \prod_{\tau = 1}^{t} \left( \frac{1}{1 + \max(0, \mathrm{CD}_{\mathrm{R}}(\hat{\set{X}}_{\tau}, \set{P}_{\tau}) - \mathrm{cd}_{\mathrm{max}})} \right)^\delta,
\end{equation}
which down-weights contributions from later frames if accumulated errors in preceding steps remain high.
Here, the exponent $ \delta $ is a schedule-dependent catch-up factor defined as:
\begin{equation}
    \delta = 1 - \sqrt{e / e_{\max}},
\end{equation}
where $ e $ denotes the current optimization epoch and $ e_{\max} $ the total number of epochs.
This ensures that $ w_{\mathrm{confidence}}(t) $ gradually increases toward $ 1 $ over time, allowing later frames to fully contribute once earlier deformations become sufficiently accurate, even for small residual errors over very long sequences.
To normalize the confidence relative to the achievable alignment, we estimate a soft upper bound on reconstruction accuracy as:
\begin{equation}
	\mathrm{cd}_{\mathrm{max}} = \max_{t \in [1:T]} \mathrm{CD}_{\mathrm{R}}(\sg(\hat{\set{P}}_t), \set{P}_t),
\end{equation}
where $ \sg(\cdot) $ denotes the stop-gradient operator, which prevents gradients from propagating through this term during backpropagation.
This prevents feedback loops that could otherwise interfere with the learning of the deformation parameters $ \vec{T}_t^{l,c} $.

\begin{figure}    
    \centering
    \includegraphics[width=\linewidth]{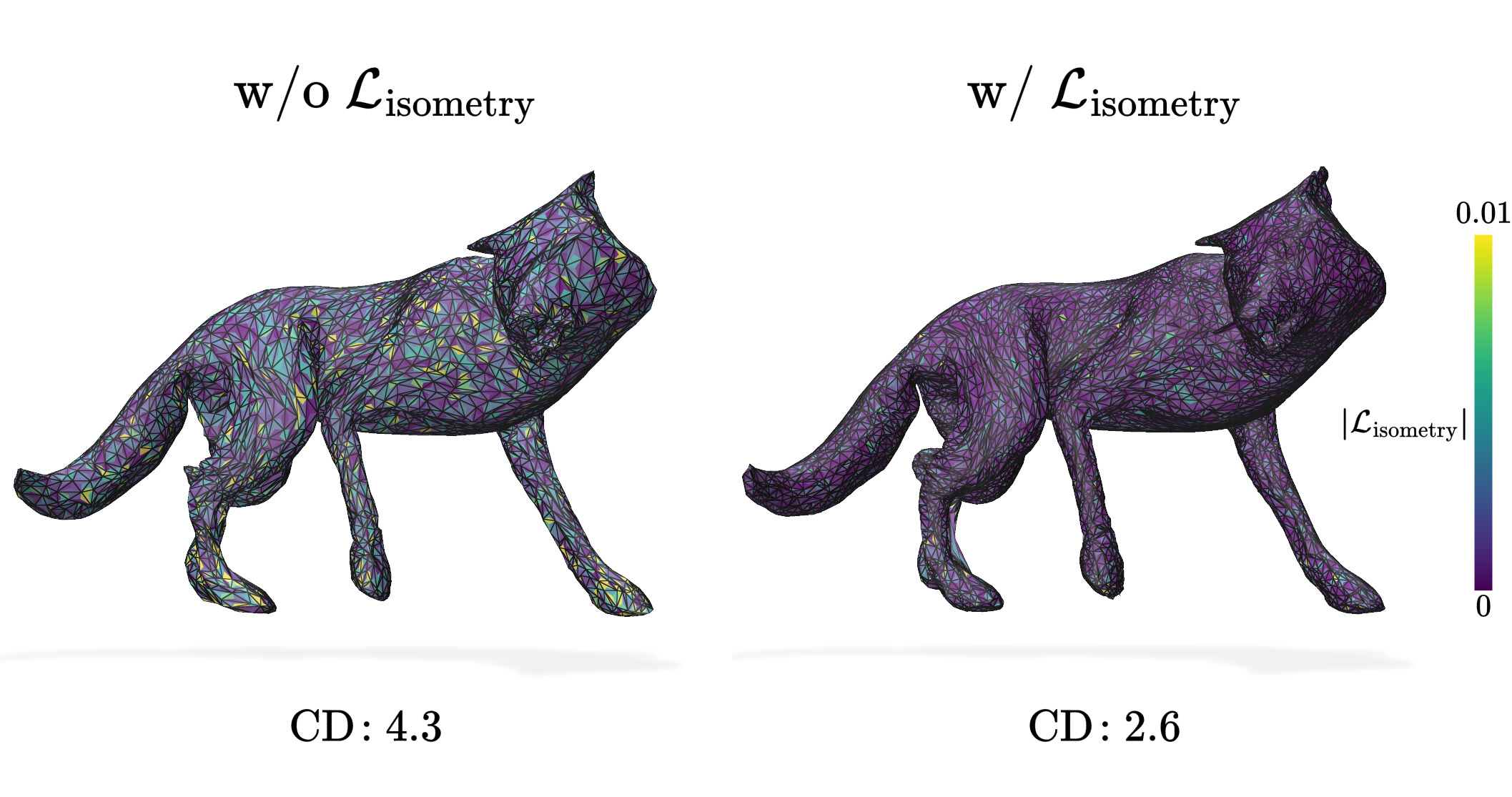}
    \caption{
    Effect of the complementary isometry loss $ \mathcal{L}_{\mathrm{isometry}} $ on the per-edge displacements.
    Without $ \mathcal{L}_{\mathrm{isometry}} $ (left), the surface is faithfully deformed but exhibits larger tangential deformations.
    When $\mathcal{L}_\mathrm{isometry}$ is applied (right), the deformations become smoother and more temporally coherent, which also leads to improved Chamfer distances (CD) $ [ \times 10^{-5}] $.
    }
    \label{fig:regularizer}
\end{figure}

\paragraph*{Isometric Regularization.}

While the surface initialization and transformation fitting losses yield smooth and stable deformations, they do not enforce constraints along the mesh surface itself, including temporal coherence in these directions.
However, in many real-world scenarios, surface motion is approximately isometric and thus intrinsic geometric properties such as edge lengths are typically preserved.
To promote such physically plausible behavior, we introduce an isometry loss that penalizes temporal variations in edge length (see \Cref{fig:regularizer}):
\begin{equation}
    \mathcal{L}_{\mathrm{isometry}} = \frac{1}{T \, \abs{\set{E}}} \sum_{t = 1}^T \sum_{(i,j) \in \set{E}} \Big\vert \norm{\hat{\vec{x}}_{i, t} - \hat{\vec{x}}_{j, t}} - \norm{\hat{\vec{x}}_{i, t - 1} - \hat{\vec{x}}_{j, t - 1}} \Big\vert,
\end{equation}
where $ \set{E} $ denotes the set of edges in the reference mesh topology.

\subsection{Keyframe Selection}
\label{sec:keyframe}

To initialize the optimization with a suitable initial surface, we follow a similar strategy as DynoSurf~\cite{yao2024dynosurf} to select a keyframe as the starting point.
Rather than choosing the frame with the lowest aggregated Chamfer distance to all others, we prioritize selecting frames that exhibit a well-defined and representative surface topology.
To this end, we measure the spatial extent of each frame and select the one with maximal coverage near the temporal midpoint:
\begin{equation}
    t_{\mathrm{key}} = \argmax_{t \in [0:T]} w_{\mathrm{key}}\left(t - \tfrac{T}{2}\right) \, \abs{\set{G}(\set{P}_t)},
\end{equation}
where $ \abs{\set{G}(\set{P}_t)} $ denotes the number of occupied voxels in a fixed-resolution grid $ \set{G} $ of size $ 128^3 $.
To favor a keyframe near the temporal center, we apply a Gaussian weighting function:
\begin{equation}
    w_{\mathrm{key}}(t) = \exp(- \gamma \, t^2), 
    \qquad 
    \gamma = 0.001,
\end{equation}
which discourages frames at the sequence boundaries, where it would be necessary to estimate larger overall transformations to other frames.
Once $ t_{\mathrm{key}} $ is determined, we reconstruct the initial mesh $ \set{X}_0 $ via screened Poisson surface reconstruction~\cite{kazhdan2013screened} applied to the corresponding point cloud $ \set{P}_{t_{\mathrm{key}}} $.

\subsection{Implementation Details}

All input point clouds $ \set{P}_t $ are normalized to the spatial domain $ [-1, 1]^3 $ to align with the expected range of our multi-resolution voxel grid representation.
To reduce computational overhead, we prune the transformation grid by retaining only cells that are either directly occupied by input points or fall within a three-cell neighborhood of occupied regions.
This sparsification strategy reduces the number of active parameters by over 50\%, substantially lowering memory consumption and accelerating optimization.
As a result, our full pipeline typically requires less than 2 GB of GPU memory and completes processing a sequence in approximately 7 minutes on an NVIDIA RTX 4090.
For optimization, we use the Adam optimizer with default hyperparameters ($ { \beta_1 = 0.9 } $, $ { \beta_2 = 0.999 } $, $ { \epsilon = 10^{-8} } $), combined with a hierarchical learning rate schedule across grid resolutions.
The coarsest grid level ($ { l = 1 } $) is optimized using a base learning rate $ \eta $ of $ { 5 \times 10^{-3} } $, which is increased by 10\% for each subsequent (finer) level.
Similarly, the preconditioning smoothness weight $ \lambda $ is initialized at $ 0.25 $ for the coarsest level and increased by 50\% per level to enforce stronger spatial coherence at higher resolutions.
The optimization of the mesh vertices of $ \set{X}_0 $ is also preconditioned, using a fixed learning rate of $ { 1 \times 10^{-4} } $ and a smoothness weight of $ { \lambda = 16 } $.

\section{Evaluation}
\label{sec:evaluation}

\begin{figure*}
    \centering
    \includegraphics[width=\textwidth]{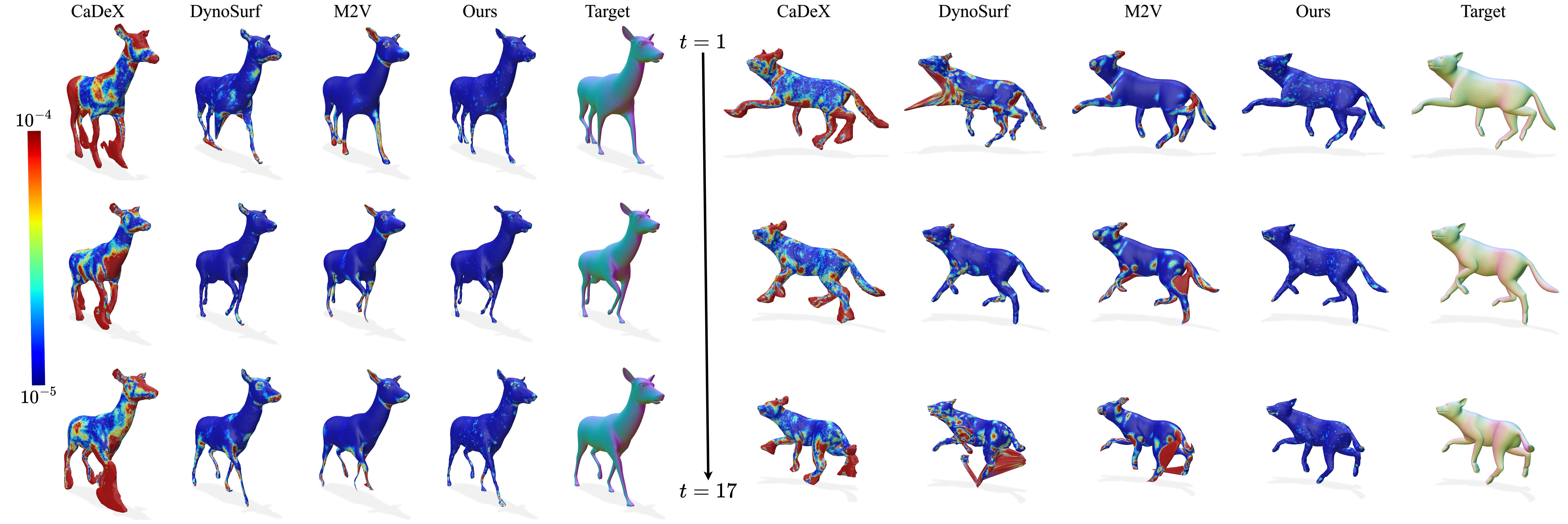}
    \caption{Comparison of visual results for CaDeX, DynoSurf, M2V, and Our method on two motion sequences of the DT4D dataset.
    Color maps indicate per-vertex $ \ell_2 $-Chamfer distance.
    Our method achieves the lowest error and best visual fidelity across all frames.}
    \label{fig:comparison}
\end{figure*}

\begin{figure*}
    \centering
    \includegraphics[width=\textwidth]{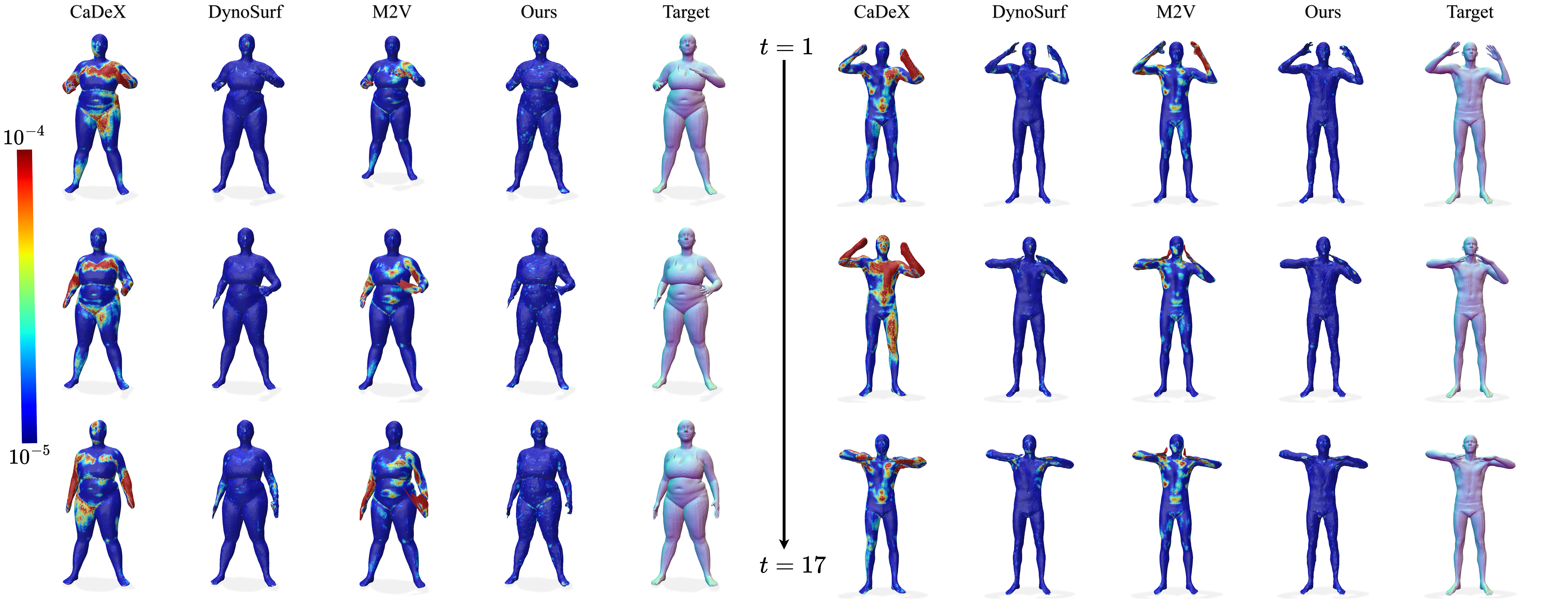}
    \caption{Comparison of visual results for CaDeX, DynoSurf, M2V, and Our method on two motion sequences of the DFAUST dataset.
    Color maps indicate per-vertex $\ell_2$-Chamfer distance.
    Our method achieves the lowest error and best visual fidelity across all frames.}
    \label{fig:comparison2}
\end{figure*}

\begin{table}
    \scriptsize
    \centering
    \caption{Quantitative comparison of reconstruction performance across common animation datasets. 
    Our method consistently achieves the lowest Chamfer distance and Correspondence Error as well as the highest Normal Consistency and F-scores, demonstrating superior accuracy and temporal coherence compared to baseline methods.}
    \setlength{\tabcolsep}{4.5pt}
    \begin{tabular}{l|l|c|ccccc}
        \toprule
        \multicolumn{1}{c}{} & & corresp. & CD $ [\times 10^{-5} ]$ $\downarrow$ & NC $\uparrow$ & F-$0.5\%$ $\uparrow$ & F-$1\%$ $\uparrow$ & Corr. $\downarrow$ \\
        \midrule
        \multirow{4}{*}{\rotatebox{90}{\tiny DFAUST}}
        & CaDeX & \xmark & \phantom{0}3.68 & 0.941 & 0.730 & 0.921 & 0.539 \\
        & DynoSurf & \xmark & \phantom{0}2.13 & 0.953 & 0.980 & 0.992 & 0.356 \\
        & M2V & \cmark & \phantom{0}1.61 &  \textbf{0.960} & 0.877 & 0.979 & 0.358 \\
        & Ours & \xmark & \phantom{0}\textbf{0.52} & 0.957 & \textbf{0.988} & \textbf{0.997} & \textbf{0.355} \\
        \midrule
        \multirow{4}{*}{\rotatebox{90}{\tiny DT4D}}
        & CaDeX & \xmark & 56.51 & 0.870 & 0.386 & 0.652 & 0.442 \\
        & DynoSurf & \xmark & 15.18 & 0.919 & 0.773 & 0.922 & \textbf{0.419} \\
        & M2V& \cmark & \phantom{0}7.61 & 0.944 & 0.792 & 0.938 & 0.425 \\
        & Ours & \xmark & \phantom{0}\textbf{1.53} & \textbf{0.960} & \textbf{0.961} & \textbf{0.994} & 0.422 \\
        \midrule
        \multirow{2}{*}{\rotatebox{90}{\tiny AMA}}
        & DynoSurf & \xmark& \phantom{0}1.01 & 0.918 & 0.921 & 0.992 & \textbf{0.347} \\
        & Ours & \xmark& \phantom{0}\textbf{0.47} & \textbf{0.939} & \textbf{0.985} & \textbf{0.999} & 0.348 \\
        \bottomrule
    \end{tabular}
    \label{tab:metrics}
\end{table}

We evaluated our method on three animation datasets encompassing both human and animal motion.
Specifically, AMA~\cite{vlasic2008articulated} and DFAUST~\cite{bogo2017dynamic} comprise diverse human motion sequences, while DT4D~\cite{li20214dcomplete} provides animal motions, offering broader coverage beyond human-centric benchmarks.
For comparisons to learning-based methods, we adopted the dataset splits of DynoSurf~\cite{yao2024dynosurf}, yielding 33 test sequences for AMA, 109 for DFAUST, and 89 for DT4D.
We used the official checkpoints and code released by each method, without additional fine-tuning, and restricted the performance evaluations to the datasets each method was originally trained on for a fair comparison.
For DynoSurf, we report both the published results for general benchmarking and results from additional experiments and ablations using the official implementation.

\subsection{Comparison to State-of-the-Art}

We compare our approach against three state-of-the-art methods for 4D surface reconstruction from point cloud sequences.
The learning-based baselines CaDeX~\cite{lei2022cadex} and Motion2VecSets (M2V)~\cite{cao2024motion2vecsets} leverage pre-trained models and incorporate category-specific priors, explicitly distinguishing between human and animal motion patterns.
In contrast, our method and DynoSurf~\cite{yao2024dynosurf} are category-agnostic and, thus, do not require pre-training or semantic priors.
Notably, M2V additionally assumes access to dense temporal point correspondences, introducing stronger requirements on input data and thereby relying on significantly more prior information.
Reconstruction quality is evaluated using $ \ell_2 $-Chamfer Distance (CD), which quantifies geometric accuracy, and Normal Consistency (NC), which measures the alignment of surface normals across reconstructions.
To assess temporal coherence, we report Correspondence Error (Corr.) and compute F-scores at 0.5\% and 1\% thresholds to quantify geometric overlap.
All meshes are normalized to a unit bounding box within $ [0, 1]^3 $ prior to evaluation.
For metric computation, 100000 points are uniformly sampled from both the predicted and ground-truth surfaces at each timestep.
Unless otherwise stated, all experiments are performed with 5000 input target points per timestep over a motion sequence consisting of 17 timesteps.
As summarized in \Cref{tab:metrics}, our method consistently outperforms baselines across all metrics and object categories, despite the absence of category-specific supervision.
Qualitative results, visualized in \Cref{fig:comparison,fig:comparison2}, further highlight the effectiveness of our method, showing Chamfer distance maps on two DT4D animal sequences and two DFAUST human sequences, respectively.

\begin{figure*}
    \centering
    \includegraphics[width=\linewidth]{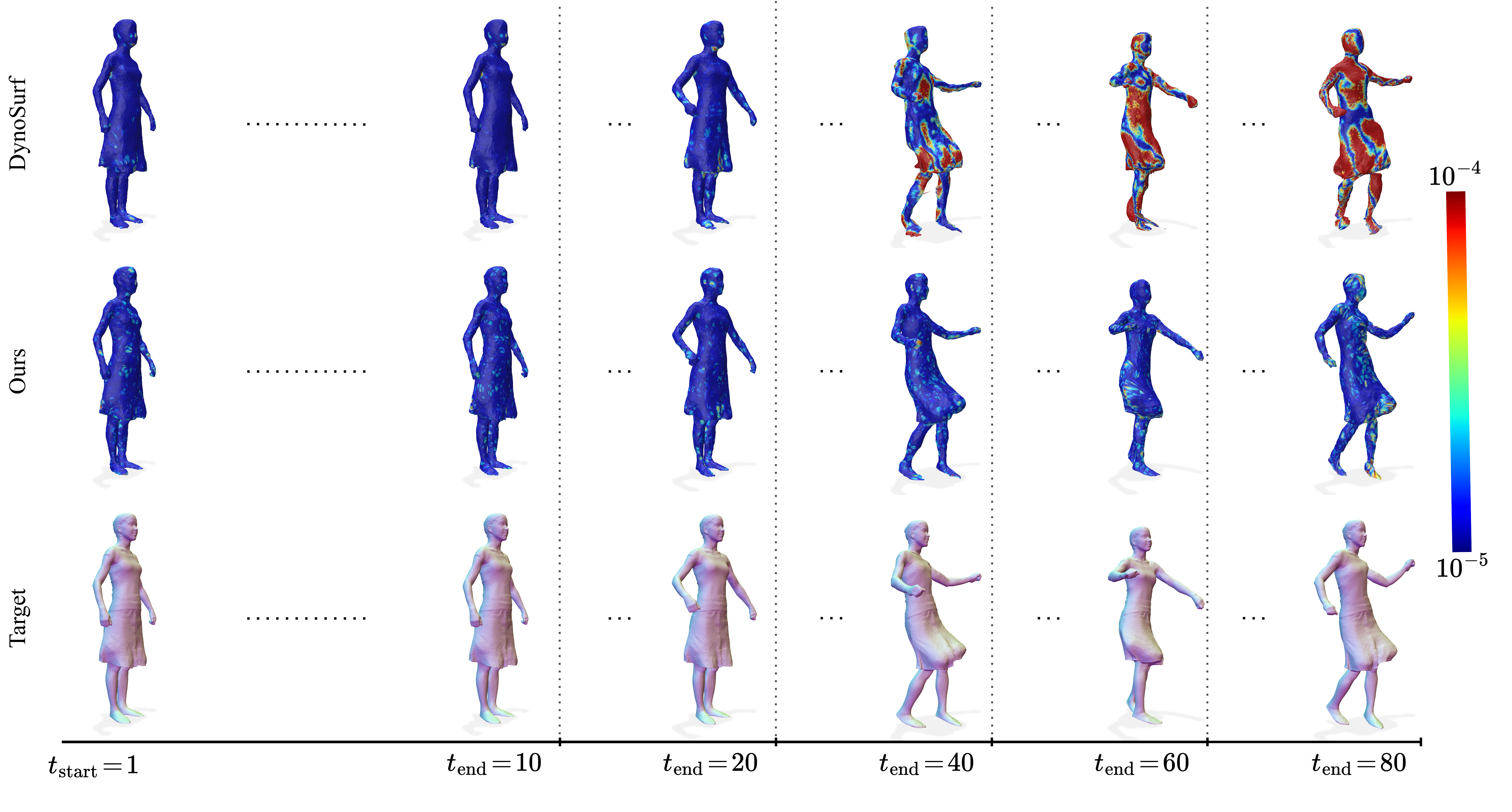}
    \caption{Comparison of CaDeX, DynoSurf, and Our method on the AMA dataset across motion sequences of increasing length, all beginning from the same initial pose. 
    The final frame of each sequence is shown, with reconstruction error visualized as $ \ell_2 $-Chamfer distance, clamped to the range $ { [10^{-5} , 10^{-4} ] } $.
    Our method consistently maintains lower error over time, demonstrating robustness to sequence length.}
    \label{fig:length}
\end{figure*}

\begin{table}
    \scriptsize
    \centering
    \caption{
    Performance across varying sequence lengths $ T $ on the AMA dataset. 
    Our method remains robust as the length increases, while DynoSurf exhibits significant performance degradation.}
    \renewcommand{\arraystretch}{1.25}
    \begin{tabular}{c|l|cccc}
        \toprule
        $ T $ & & CD $ [\times 10^{-5} ]$ $\downarrow$ & NC $\uparrow$  & F-$0.5\%$ $\uparrow$  & F-$1\%$ $\uparrow$ \\
        \midrule
        \multirow{2}{*}{{10}}
        & DynoSurf & \phantom{0}0.80 &0.919 & 0.943 & 0.996 \\
        & Ours & \phantom{0}0.55 & 0.928 & 0.980 & 0.999  \\
        \midrule
        \multirow{2}{*}{{20}}
        & DynoSurf & \phantom{0}2.22 & 0.886 & 0.809 & 0.961 \\
        & Ours & \phantom{0}0.54 & 0.928 & 0.979 & 0.999 \\
        \midrule
        \multirow{2}{*}{{40}}
        & DynoSurf & \phantom{0}4.64 & 0.857 & 0.686 & 0.902 \\
        & Ours & \phantom{0}0.66 & 0.923 & 0.971 & 0.999 \\
        \midrule
        \multirow{2}{*}{{60}}
        & DynoSurf & 14.64 & 0.778 & 0.474 & 0.743 \\
        & Ours & \phantom{0}0.72 & 0.919 & 0.960 & 0.997 \\
        \midrule
        \multirow{2}{*}{{80}}
        & DynoSurf & 16.32 & 0.759 & 0.431 & 0.700 \\
        & Ours & \phantom{0}1.35 & 0.906 & 0.931 & 0.990 \\
        \bottomrule
    \end{tabular}
    \label{tab:length}
\end{table}

\paragraph*{Sequence Length Analysis.}

The accuracy of transformation estimation can vary significantly with the motion sequence length.
To analyze this dependency, we evaluated the performance of our method across sequences of varying lengths and compared it to alternative approaches that do not employ correlation-based mechanisms.
This study is conducted on the AMA~\cite{vlasic2008articulated} dataset, which features complex, long-range human motions well-suited for assessing temporal robustness.
For the analysis, the sequences are partitioned into fixed lengths of 10, 20, 40, 60, and 80 time frames.
As shown in \Cref{fig:length} and \Cref{tab:length}, we visualize the reconstructions at the final timestep of each segment, illustrating our method’s capability to capture extended temporal deformations with high accuracy.
For longer sequences, we proportionally increase the number of optimization steps up to 10000 to account for our method’s frame-by-frame optimization strategy described in \Cref{sec:transform_param}.

\subsection{Ablation Study}

To assess the contribution of each component in our pipeline, we conducted a series of ablation studies on the test split of the AMA~\cite{vlasic2008articulated} dataset.
In particular, we examined the method’s sensitivity to input noise, the effect of varying the number of levels in the multi-resolution grid, and the impact of key components like smoothness preconditioning and the isometry loss.
These studies offer insights into how specific design choices influence reconstruction accuracy, robustness to noise, and optimization stability.

\begin{figure*}
    \centering
    \includegraphics[width=\textwidth]{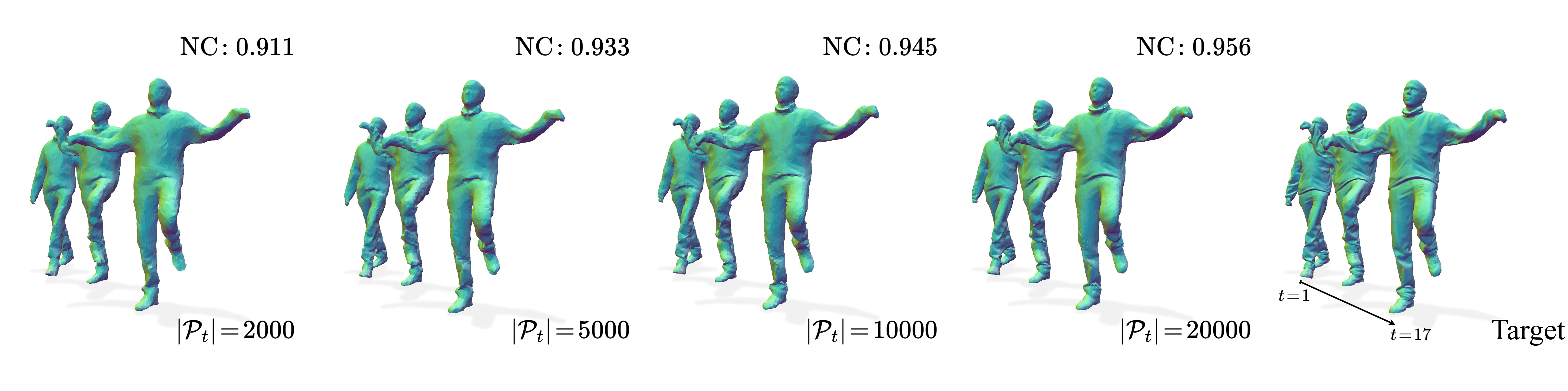}
    \caption{Effect of point cloud resolution on reconstruction quality. 
    Our method remains robust across varying numbers of input points, maintaining high accuracy even at lower resolutions. }
    \label{fig:resolution}
\end{figure*}

\paragraph*{Point Cloud Resolution.}

For methods that do not leverage pretrained knowledge of the transformed surfaces, the resolution of the target point clouds plays a critical role in reconstruction performance.
Increasing the number of target points can also help expose the performance ceiling of such approaches.
To investigate this, we evaluated each method using target point clouds ranging from 2500 to 20000 points.
As shown in \Cref{tab:metrics_res}, our method scales effectively with resolution, while alternative methods exhibit early saturation in performance at lower resolutions.
In addition, surface normals and object-specific normal consistency for the reconstructed geometries are visualized in \Cref{fig:resolution}.

\begin{table}
    \scriptsize
    \centering
    \caption{
    Performance at varying point cloud resolutions. 
    Our method consistently improves with higher input densities, while DynoSurf shows limited scalability.}
    \renewcommand{\arraystretch}{1.25}
    \begin{tabular}{c|l|cccc}
        \toprule
        $ \abs{\set{P}_t} $ & & CD $ [\times 10^{-5} ] $ $\downarrow$ & NC $\uparrow$  & F-$0.5\%$ $\uparrow$  & F-$1\%$ $\uparrow$ \\
        \midrule
        \multirow{2}{*}{{\phantom{0}2500}}
        & DynoSurf & 1.56 & 0.896 & 0.862 & 0.981 \\
        & Ours & 0.79 & 0.922 & 0.948 & 0.997  \\
        \midrule
        \multirow{2}{*}{{\phantom{0}5000}}
        & DynoSurf & 1.01 & 0.918 & 0.921 & 0.992 \\
        & Ours & 0.47 & 0.939 & 0.985 & 0.999 \\
        \midrule
        \multirow{2}{*}{{10000}}
        & DynoSurf & 1.28 & 0.906 & 0.897 & 0.986 \\
        & Ours & 0.40 & 0.950 & 0.993 & 0.999 \\
        \midrule
        \multirow{2}{*}{{20000}}
        & DynoSurf & 1.45 & 0.902 & 0.887 & 0.982 \\
        & Ours & 0.37 & 0.959 & 0.996 & 0.999 \\
        \bottomrule
    \end{tabular}
    \label{tab:metrics_res}
\end{table}

\paragraph*{Key Components.}
We conducted an ablation study to evaluate the contributions of three key components of our grid optimization: the multi-resolution grid structure, smoothness preconditioning of the grid cells, and the isometry loss.
To ensure a fair comparison, we reduced the grid learning rate to 10\% of its original value when smoothness preconditioning is disabled, as this component is essential for stable optimization at higher step sizes.
As shown in \Cref{tab:structure}, only enabling the multi-resolution grid already leads to the most substantial performance improvement as it allows to model deformations across multiple spatial scales.
Adding smoothness preconditioning further enhances reconstruction quality by encouraging stable, coherent transformation updates during optimization. 
Its impact is most apparent when combined with the multi-resolution grid, particularly in the surface metrics, normal consistency and F-scores, by promoting smooth motion of neighboring regions and thereby improving surface reconstruction quality.
While the isometry loss contributes less in terms of quantitative metrics, it plays an important role in constraining local surface distortions and preserving temporal consistency, particularly in near-static or under-constrained regions.

\begin{table}
    \scriptsize
    \centering
    \caption{
    Ablation on key components of our method. 
    When the multi-resolution voxel grid is disabled, only the finest resolution level is used. 
    Additionally, in the absence of preconditioning, the learning rate of the grid is reduced to 10\%. 
    Results show that the multi-resolution grid alone yields the largest gain in Chamfer distance, while preconditioning improves F-scores and normal consistency. 
    The full model achieves the best overall performance.}
    \setlength{\tabcolsep}{4.0pt}
    \begin{tabular}{c c c|cccc}
        \toprule
        Smooth. Prec. & Multi-Res. & Isometry & CD $ [\times{10^{-5}} ] $ $\downarrow$ & NC $\uparrow$ & F-$0.5\%$ $\uparrow$ & F-$1\%$ $\uparrow$ \\
        \midrule
        - & - & - & 9 $\times 10^5$ & 0.796 & 0.720 & 0.791 \\
        - & - & \cmark & 3$\times 10^2$ & 0.901 & 0.929 & 0.967  \\
        - & \cmark & - & 0.69 & 0.910 & 0.960 & 0.997 \\
        - & \cmark & \cmark & 4.39 & 0.913 & 0.958 & 0.991  \\
        \cmark & - & - & 2.54 & 0.901 & 0.950 & 0.991 \\
        \cmark & - & \cmark & 4.37 & 0.913 & 0.958 & 0.991  \\
        \cmark & \cmark & - & 0.67 & 0.933 & 0.978 & 0.998  \\
        \cmark & \cmark & \cmark & 0.47 & 0.939 & 0.985 & 0.999  \\
        \bottomrule
    \end{tabular}
    \label{tab:structure}
\end{table}

\begin{table}
    \scriptsize
    \centering
    \caption{Reconstruction performance under varying levels of Gaussian noise added to the input point clouds.
    The noise magnitude is expressed as a percentage of the bounding box diagonal.
    Results demonstrate our method’s robustness, with stable reconstruction quality observed across moderate noise levels.}
    \renewcommand{\arraystretch}{1.25}
    \begin{tabular}{c|cccc}
        \toprule
        Noise &  CD $ [\times 10^{-5} ]$ $\downarrow$ & NC $\uparrow$  & F-$0.5\%$ $\uparrow$  & F-$1\%$ $\uparrow$ \\
        \midrule
        0\%  & \phantom{0}0.47 & 0.939 & 0.985 & 0.999  \\
        0.25\% & \phantom{0}0.76 & 0.912 & 0.958 & 0.998 \\
        0.5\% & \phantom{0}1.25 & 0.865 & 0.870 & 0.995 \\
        1\% & \phantom{0}3.83 & 0.703 & 0.581 & 0.902 \\
        2\% & 20.90 & 0.529 &0.254 & 0.508 \\
        \bottomrule
    \end{tabular}
    \label{tab:noise}
\end{table}

\paragraph*{Noise.}

We evaluated the robustness of our method to input noise by adding Gaussian noise to the point clouds.
The noise magnitude is chosen based on the bounding box diagonal of the input point cloud, with levels set to 0.25\%, 0.5\%, 1\%, and 2\%.
As shown in \Cref{tab:noise}, our method maintains high reconstruction quality under moderate noise, despite not employing any explicit denoising strategy.
While performance degrades gradually with increasing noise, the results remain stable up to 1\%, demonstrating the strong resilience of our method to imperfect input.
Significant degradation is observed only at the extreme noise level of 2\% which, however, represents an unlikely real-world use case.

\begin{figure*}
    \centering
    \includegraphics[width=\textwidth]{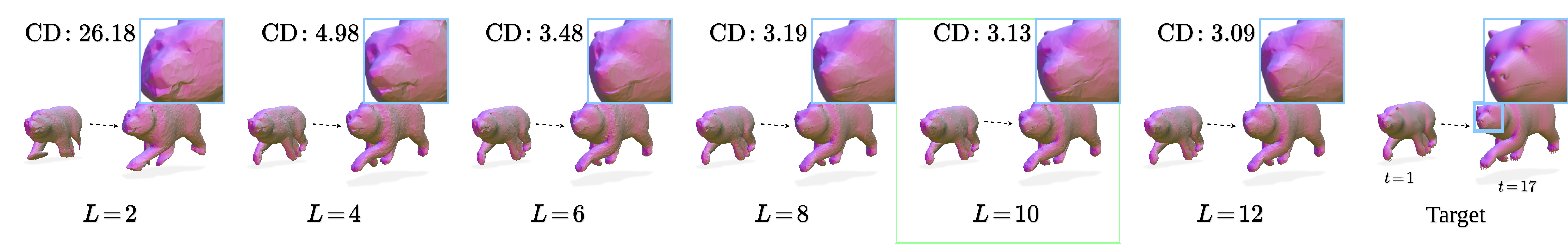}
    \caption{Effect of the number of grid levels on reconstruction quality. We report the Chamfer distance (CD) $ [10^{-5}] $ to highlight reconstruction accuracy. 
    Our default configuration with 10 levels is marked in green, while zoom-ins on key regions are shown in blue. 
    Fewer levels already yield reasonable results, but additional levels enhance fine-scale detail without compromising stability.}
    \label{fig:grid_level}
\end{figure*}

\paragraph*{Grid Size.}

We also investigated the impact of the grid resolution by varying the number of levels in our hierarchical voxel structure.
Specifically, we compare configurations with 1, 2, 4, 6, 8, and 12 levels against our default setup with 10 levels.
As illustrated in \Cref{fig:grid_level}, even low-resolution configurations with significantly fewer grid cells are capable of providing high-quality reconstructions.
Increasing the number of levels consistently improves the reconstruction accuracy by capturing finer-scale local deformations.
However, gains beyond 8 levels become increasingly marginal, while the default configuration of 10 levels strikes a good balance between accuracy and computational cost.
Crucially, the added flexibility of higher resolutions (12 levels) does not degrade performance, demonstrating the stability and scalability of our approach.

\begin{figure}
    \centering
    \includegraphics[width=\linewidth]{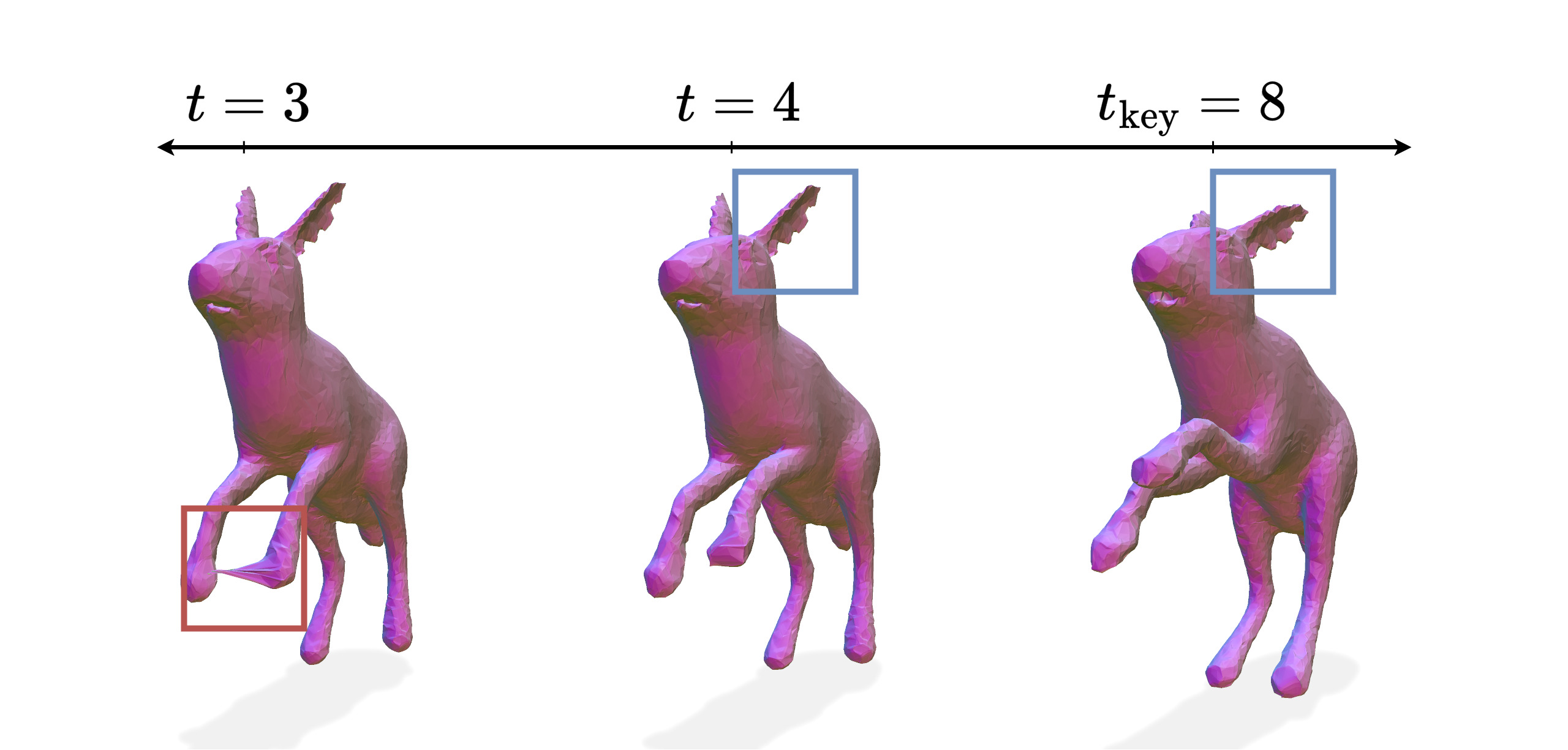}
    \caption{Failure cases of our method. 
    Blue regions highlight artifacts caused by sparse input sampling at the initial timestep, leading to errors in the initial surface reconstruction. 
    Red regions indicate correspondence errors resulting from insufficient alignment of the input points during grid optimization.}
    \label{fig:limitation}
\end{figure}

\paragraph*{Initialization.}

We evaluated how keyframe selection $ t_{\mathrm{key}} $ and surface initialization affect reconstruction quality.
Specifially, we compared 1) choosing the first frame, 2) the temporal middle frame, and 3) our coverage-weighted keyframe~(see \Cref{sec:keyframe}), as well as 1) screened Poisson surface reconstruction~\cite{kazhdan2013screened}, 2) the deformable tetrahedron of DynoSurf~\cite{yao2024dynosurf}, and 3) a pretrained diffusion initializer~\cite{cao2024motion2vecsets}.
As shown in \Cref{tab:init}, Poisson reconstruction consistently achieved robust results on the DT4D dataset and is insensitive to the keyframe choice.
The diffusion initializer may achieve higher geometric accuracy when seeded with a favorable keyframe, but it highly relies on good keyframes $ t_{\mathrm{key}} $ and quickly degrades otherwise.
In general, both the middle and our selection scheme yield small, yet reliable improvements over picking the first frame.
Considering the AMA dataset, where diffusion would require retraining and is therefore excluded, Poisson reconstruction also outperforms the tetrahedron baseline.
Similarly, choosing the middle frame or our coverage-based keyframe consistently improves the metrics compared to the first frame.
Here, our scheme typically matches the middle frame very closely in practice, making it a reliable proxy for the unknown optimal frame.
Especially for the tetrahedron initialization, this leads to substantial gains and narrows down the gap to Poisson reconstruction.
Overall, Poisson reconstruction offered the best trade-off between accuracy and stability, especially when paired with our keyframe selection.

\begin{table}
    \scriptsize
    \centering
    \caption{We compare the behavior of our method when selecting the first or middle frame as the keyframe, against our proposed keyframe selection strategy. In addition, we evaluate different point cloud-to–surface initialization methods, comparing our screened Poisson reconstruction to the deformable tetrahedron approach~\cite{yao2024dynosurf} and the pretrained diffusion model~\cite{cao2024motion2vecsets}.}
    \begin{tabular}{c|c|l|cccc}
        \toprule
         \multicolumn{1}{c}{} & $t_{\mathrm{key}}$ & Method & CD $ [\times 10^{-5} ] $ $\downarrow$ & NC $\uparrow$  & F-$0.5\%$ $\uparrow$  & F-$1\%$ $\uparrow$ \\
        \midrule
        \multirow{9}{*}{\rotatebox{90}{DT4D}}
        & \multirow{3}{*}{{First}}
        & Diffusion & 8.42 &0.956 & 0.942 & 0.984 \\
       &&  Tetrahedron  & 5.07 & 0.912 & 0.879 & 0.940 \\
        &&  Poisson & 2.50 & 0.959 & 0.951 & 0.992  \\
        \cmidrule{2-7}
        & \multirow{3}{*}{{Middle}}
        & Diffusion & 1.87 & 0.962 & 0.956 & 0.992 \\
        && Tetrahedron  & 6.70 & 0.925 & 0.902 & 0.948 \\
        && Poisson & 2.32 & 0.962 & 0.961 & 0.994  \\
        \cmidrule{2-7}
        & \multirow{3}{*}{{Ours}}
        & Diffusion & 3.53 & 0.960 & 0.954 & 0.989 \\
        && Tetrahedron  & 4.45 & 0.925 & 0.905 & 0.954 \\
        && Poisson & 2.35 & 0.962 & 0.961 & 0.994  \\
        \midrule
        \multirow{6}{*}{\rotatebox{90}{AMA}}
        &\multirow{2}{*}{{First}}
        & Tetrahedron  & 0.60 & 0.930 & 0.973 & 0.998 \\
        && Poisson & 0.59 & 0.931 & 0.975 & 0.999  \\
        \cmidrule{2-7}
        &\multirow{2}{*}{{Middle}}
        & Tetrahedron  & 0.60 & 0.933 & 0.977 & 0.998 \\
        && Poisson & 0.52 & 0.935 & 0.981 & 0.999  \\
        \cmidrule{2-7}
        &\multirow{2}{*}{{Ours}}
        & Tetrahedron  & 0.54 & 0.934 & 0.978 & 0.999 \\
        && Poisson & 0.53 & 0.935 & 0.981 & 0.999  \\
        \bottomrule
    \end{tabular}
    \label{tab:init}
\end{table}

\subsection{Limitations}

While our method demonstrates strong robustness across a range of scenarios, certain failure cases remain, as illustrated in \Cref{fig:limitation}. 
One limitation arises when the point sampling at the keyframe is too sparse. 
In this case, the initial surface reconstruction based on Laplacian regularization may exhibit artifacts that persist throughout the sequence. 
This occurs because the surface mesh is optimized solely to match the sparse keyframe point cloud, which may lack sufficient detail to constrain the geometry accurately. 
The transformation grid cannot resolve these artifacts either, as the corresponding erroneous regions are represented by too few points in subsequent timesteps to trigger corrective deformations.
Another limitation involves occasional errors in transformation estimation, particularly when the grid is not sufficiently optimized with respect to the input. 
This can lead to inaccurate local correspondences, which propagate over time and degrade alignment quality. 
These issues are partially influenced by the confidence scaling term, which controls the influence of previous steps in the optimization.

\section{Conclusion}

We introduced Preconditioned Deformation Grids, a correspondence-free and training-free technique for estimating coherent deformation fields directly from unstructured point cloud sequences.
Our method addressed the inherently under-constrained nature of this problem by employing Sobolev preconditioning, which spatially diffuses gradient information to achieve a spatially adaptive smoothness.
We further guided the optimization using multi-resolution voxel grids to represent the deformation field, allowing coarser levels to maintain temporal coherence over long sequences and finer levels to capture high-frequency surface details.
Through extensive qualitative and quantitative experiments, we demonstrated that our method achieves superior reconstruction results over existing methods using only a simple Chamfer loss and a weak isometry loss, providing a robust and flexible solution for arbitrary object motion without relying on restrictive priors or extensive training data.
\paragraph*{Future Work.}
Our framework offers a solid foundation for several natural extensions. 
Beyond point clouds, adapting it to richer representations such as 3D Gaussian Splatting or implicit neural primitives could broaden its applicability to 4D capturing scenarios. 
Moreover, modifying the grid structure toward physically motivated formulations may enable the estimation of complex dynamics including fluid motion, thereby bringing reconstruction and simulation closer together. 
Finally, exploring adaptive or learned preconditioning strategies could further improve robustness by adjusting smoothness to local data characteristics.

\subsection*{Acknowledgements}
This research has been funded by the Federal Ministry of Education and Research under grant no. 01IS22094A WEST-AI, 
by the Federal Ministry of Education and Research of Germany as well as the state of North-Rhine Westphalia as part of the Lamarr-Institute for Machine
Learning and Artificial Intelligence,
by the Ministry of Culture and Science North Rhine-Westphalia under grant number PB22-063A (InVirtuo 4.0: Experimental Research in Virtual Environments) 
and by the state of North Rhine-Westphalia as part of the Excellency Start-up Center.NRW (U-BO-GROW) under grant number 03ESCNW18B.
Open Access funding enabled and organized by Projekt DEAL.

\bibliographystyle{eg-alpha-doi} 
\bibliography{main}


\end{document}